\documentclass{article}

\usepackage[nonatbib,preprint]{neurips_2023}

\usepackage[utf8]{inputenc} 
\usepackage[T1]{fontenc}    
\usepackage{hyperref}       
\usepackage{url}            
\usepackage{booktabs}       
\usepackage{amsfonts}       
\usepackage{nicefrac}       
\usepackage{microtype}      
\usepackage{xcolor}         
\usepackage{multirow}

\usepackage{amsmath}
\usepackage{amssymb}
\usepackage{mathtools}
\usepackage{amsthm}

\usepackage{amsfonts}       
\usepackage{nicefrac}       
\usepackage{microtype}      
\usepackage{comment}
\usepackage{enumitem}

\usepackage{colortbl}
\usepackage{graphicx,color}
\usepackage{clrscode}
\usepackage{subfigure}
\usepackage{adjustbox}
\usepackage{multirow}
\usepackage{longtable}
\usepackage{clrscode}
\usepackage{array}
\usepackage{wrapfig}
\usepackage{enumitem}

\newcounter{noteZZctr} 

\usepackage{array}
\usepackage{arydshln}
\usepackage{titlesec}
\titlespacing*{\section}{0pt}{0.1\baselineskip}{0.1\baselineskip}
\titlespacing*{\subsection}{0pt}{0.1\baselineskip}{0.1\baselineskip}
\titlespacing*{\subsubsection}{0pt}{0.1\baselineskip}{0.1\baselineskip}

\title{Large Language Models for Spatial Trajectory Patterns Mining}

\author{\parbox{0.9\linewidth}{
\centering{ Zheng Zhang, Hossein Amiri, Zhenke Liu, Andreas Züfle, Liang Zhao 
} \\
{\rm Department of Computer Science\\
   Emory University \\
   Atlanta, GA 30322, USA }\\
 \texttt{\{zheng.zhang,hossein.amiri,zhenke.liu,azufle,liang.zhao\}@emory.edu} \\
}
}

\begin{document}

\maketitle

\begin{abstract}
  Identifying anomalous human spatial trajectory patterns can indicate dynamic changes in mobility behavior with applications in domains like infectious disease monitoring and elderly care. Recent advancements in large language models (LLMs) have demonstrated their ability to reason in a manner akin to humans. This presents significant potential for analyzing temporal patterns in human mobility. In this paper, we conduct empirical studies to assess the capabilities of leading LLMs like GPT-4 and Claude-2 in detecting anomalous behaviors from mobility data, by comparing to specialized methods. Our key findings demonstrate that LLMs can attain reasonable anomaly detection performance even without any specific cues. In addition, providing contextual clues about potential irregularities could further enhances their prediction efficacy. Moreover, LLMs can provide reasonable explanations for their judgments, thereby improving transparency. Our work provides insights on the strengths and limitations of LLMs for human spatial trajectory analysis.
\end{abstract}

\section{Introduction}
The widespread adoption of location-enabled mobile devices has led to a massive collection of human mobility data~\cite{mokbel2023towards}, comprising diverse trajectory types from individual app usages to public transportation systems. These mobility traces can be modeled as dynamic graphs, representing sequences of location visits with associated semantics~\cite{parent2013semantic}. Analyzing these dynamic graphs enables valuable insights for applications like transportation mode classification and detecting spatiotemporal patterns~\cite{lee2007trajectory,yao2017trajectory, biljecki2013transportation,zheng2010understanding,toohey2015trajectory,zhang2021representation,bai2023staleness,guo2022graph}. A particularly difficult task is identifying anomalous mobility patterns within an individual's semantic trajectories, where the trajectory significantly deviates from their historical patterns. Finding such anomalous patterns of individuals may indicate a change in behavior which has many important applications. For instance, infectious disease monitoring~\cite{mokbel2020contact,RambhatlaZSSL22, zhang2022unsupervised} or tracking elderly behaviors~\cite{stavropoulos2020iot}.

In recent times, there has been a surge in progress with large language models (LLMs)~\cite{qiu2020pre, ling2023domain} like Transformers~\cite{Vaswani2017AttentionIA}, BERT~\cite{devlin-etal-2019-bert}, GPT~\cite{Brown2020LanguageMA}, among others. These LLMs act as foundational models, which can be easily adapted for various downstream applications with minimal adjustments~\cite{Brown2020LanguageMA,kojima2022large,ling2023domain,zhao2023survey}. 
Notably, breakthroughs in design and training techniques have enabled emerging abilities in LLMs, distinguishing cutting-edge models like GPT-3.5~\cite{Brown2020LanguageMA}, GPT-4~\cite{OpenAI2023GPT4TR}, Claude-2~\cite{bai2022training}, BARD~\cite{bard2022}, LlaMA~{\cite{Touvron2023LLaMAOA}}, and LlaMA-2~{\cite{Touvron2023Llama2O}} from earlier versions. For example, features such as in-context training~\cite{min2021metaicl} and zero-shot learning~\cite{kojima2022large, wei2021finetuned} allow these models to adapt to tasks they were not explicitly trained for. 

Despite the remarkable progress LLMs have made in diverse NLP tasks like question answering (QA) and machine translation, their potential in analyzing human mobility patterns remains largely unexplored. Human mobility data, unlike typical language sequences, presents with intricate spatial-temporal dynamics and rich topological connections between entities. Detecting anomalous behaviors are especially difficult due to the intrinsic property of unkown nature of anomalies. Existing methods typically rely on creating hand-crafted features such as the total traveled distance and use heuristic rule to determine outliers, which limits their capability to generalize effectively to detect unseen outlier patterns. In contrast, LLMs has natural advanatge since they can directly perceive natural langauge input. As LLMs have already shown powerful reasoning ability and generalization capabilities directly from the input prompt, it becomes intriguing to assess to what extent LLMs can detect diverse anomaly behaviors under the human mobility patterns.

To systematically study the capabilities of LLMs on detecting outliers (anomlies) in human mobility trajectories, we conduct a series of empirical experiments with leading LLMs on diverse datasets. By comparing their performance to specialized human mobility anomaly detection methods, we aim to assess the potential strengths and limitations of LLMs in this domain. Critically, by altering the input prompt formats, we aim to evaluate how effectively LLMs can extract and leverage the underlying structural information from the dynamic mobility patterns to enhance their performance in subsequent tasks. Moreover, we delve into both the effectiveness and interpretability of LLMs' predictions.

\section{Related Works}
In recent literature, a few preliminary studies~\cite{xue2022leveraging, wang2023would, musleh2022let} have made attempts to uncover the potential of applying LLMs in analyzing human mobility patterns. Xue et al.~\cite{xue2022leveraging} propose a pipeline called AuxMobLCast that leverages pre-trained language models for human mobility forecasting by transforming numeric time series data into natural language descriptions. Experiments on real-world mobility datasets demonstrate that fine-tuning language models like BERT with mobility prompts can effectively capture sequential patterns and yield good performance for predicting future visitor numbers. Wang et al.~\cite{wang2023would} propose LLM-Mob that leverages LLMs for human mobility prediction by formatting mobility data into historical and context stays and designing effective prompts. Their experiments the potential of harnessing LLMs for mobility prediction through careful instructional prompting. Musleh et al.~\cite{musleh2022let} envision a BERT-like system for trajectory analysis but note challenges like the high number of distinct GPS points compared to words, noisy trajectory data, and long unrelated trajectories, necessitating customization of BERT for trajectories rather than direct application. Although some initial studies exist, our research is the first to comprehensively examine the capability of LLMs in identifying anomalies in human mobility patterns and compare them with state-of-the-art human mobility anomaly detection algorithms.

\section{Experiments}

\subsection{Experimental Settings}

\paragraph{Datasets.} We conducted the experiments on two human mobility benchmark datasets: \textsc{GeoLife}~\cite{zheng2010geolife} and \textsc{Patterns-of-Life}~\cite{zufle2023urban,kim2020location}. Brief descriptions of the datasets are as follows:

\begin{table}[t] \centering
\begin{tabular}{p{1.5cm}p{11.5cm}}\toprule
Task \newline setting & Prompt to LLM\\ \midrule
\textbf{Separate} &  Task: You are a human mobility trajectory behavior anomaly detector. Given a historical human trajectory information, can you analyse the pattern behind the trajectory and give an anomaly score (from 0 to 1, where larger value indicates more abnormal) of this user's behavior? \newline Here is the sequence of trajector: \textcolor{blue}{<Sequence>}. \newline Give your analysis and present your esimated anomaly score (from 0 to 1, where larger value indicates more abnormal) inside a pair of square brackets. \\ \hline
\textbf{Separate With Hint} &  Task: (\dots Same as Above \dots) \newline Hint: The anomaly users would suddenly change their mobility pattern starting from a time point, which means after a certain time, their mobility behavior would significantly deviate from their past behaviors. We would use ``***<deviate-point>***'' inside each trajectory to denote the time point as hint.\newline Here is the sequence of trajector: \textcolor{blue}{<Sequence-before>} ***<deviate-point>*** \textcolor{blue}{<Sequence-after>}. \newline Give your analysis and present your esimated anomaly score (from 0 to 1, where larger value indicates more abnormal) inside a pair of square brackets. \\ \hline
\textbf{Combine} & Task: You are a human mobility trajectory behavior anomaly detector. Given a set of N users' historical human trajectories information, can you analyse the pattern behind each user's trajectory and give an anomaly score (from 0 to 1, where larger value indicates more abnormal) of users' behavior?\newline
Here is the sequence of user 1: \textcolor{blue}{<Sequence-1>}\newline
\dots \newline
Here is the sequence of user N: \textcolor{blue}{<Sequence-N>}\newline
Give your analysis and present your esimated anomaly scores about all users (from 0 to 1, where larger value indicates more abnormal):\\
\bottomrule
\end{tabular}
\caption{Examples of different prompts used in anomaly detection experiments. We also have `Combine With Hint' prompt which is the similar way of adding hints in `Separate With Hint' prompt. A detailed prompt example is given in Appendix~\ref{sec:prompt} due to space limitation.}
\label{prompt}
\vspace{-5mm}
\end{table}
\begin{itemize}[leftmargin=*]

    \item \textsc{GeoLife}: This dataset was created using the Microsoft Research Asia's GPS Trajectory dataset~\cite{zheng2010geolife}. The GeoLife dataset, sourced from the GeoLife project by Microsoft Research Asia, captures the GPS trajectories of 182 users over a span of more than three years, from April 2007 to August 2012. Each trajectory in this dataset consists of time-stamped points detailing the latitude, and  longitude. The dataset provides insights into leisure and sports activities, such as shopping, sightseeing, dining, hiking, and cycling, offering a comprehensive view of users' outdoor movements. We first eliminated agents with fewer than 50 records, resulting in a final count of 69 users. Due to the lack of ground truth anomalies, we introduced a specific outlier type called the ``imposter outlier", by switching the trajectories with another agent after a specific time point. In this paper we choose 80\% of the stay points of trajectories as the switching point.
    \item \textsc{Patterns-of-Life (PoL)}: A simulated dataset, where agents emulate human activities such as working, socializing, and more, in a real-world-like setting sourced from OpenStreetMap~\cite{bennett2010openstreetmap}. Throughout their simulated lives, agents navigate to diverse locations, including restaurants, workplaces, residential apartments, and recreational venues. We generate 1,000 users over a span of 4 weeks. We introduced 90 anomaly users with three specific types of abnormal behavior:
    (1) \textbf{Hunger outlier:} An agent under this category becomes hungry more quickly. Such agents have to go to restaurants or their homes much more often.
    (2) \textbf{Social outlier:} This type of agent randomly selects recreational sites to visit when needed, rather than being guided by their attributes and social network.
    (3) \textbf{Work outlier:} Agents in this category abstain from going to work on workdays.
\end{itemize}


\paragraph{Comparison Methods.} We compare with several unsupervised trajectory outlier detection methods, including three non-deep learning methods \textbf{OMPAD} \cite{basharat2008learning}, \textbf{MoNav-TT }\cite{zhang2012smarter} and \textbf{TRAOD } \cite{lee2008trajectory}, and two state-of-the-art deep learning methods \textbf{DSVDD} \cite{ruff2018deep} and \textbf{DAE} \cite{zhou2017anomaly,dotti2020hierarchical}. A detailed introduction of the comparison methods can be found in Appendix~\ref{sec:compare}.

\subsection{LLM Detection Results}
Broadly, this paper focuses on studying the central question of investigating the capabilities of LLMs on identifying anomalous behaviors within human mobility patterns from three perspectives:
\begin{itemize}[leftmargin=*]
    \item \textbf{Can LLMs effectively detect anomalous behaviors within human mobility patterns without any indicative information?} It is intriguing to assess whether LLMs can attain substantial predictive performance on anomaly detection tasks, even in the absence of any clue about the anomalies, e.g. such as temporal occurrence or the nature of the anomaly.
    \item \textbf{Can providing indicative clues about the anomaly enhance the detection efficacy of LLMs?} Incorporating specific clues or hints about potential anomalies might bolster the LLM's ability to identify irregularities more accurately. By offering contextual information, it could guide the LLM to focus on certain aspects of the data and make more informed predictions.
    \item \textbf{Can LLMs provide reasonable explanation to their judgements?} Beyond mere classification, it is imperative to observe whether LLMs can elucidate the fundamental reasoning behind their determinations. Specifically, can these models articulate the underlying rationale when predicting human mobility patterns as anomalous or normal, thereby enhancing the transparency and trustworthiness of their judgments?
\end{itemize}\vspace{-3mm}

\paragraph{Prompt Settings.} To systematically study the research questions above, we design two different dimensions to create input prompts for the LLMs. Specifically, (1) \textbf{With/Without Hint}: Given that an anomaly begins at a specific time point in the data, it is crucial to evaluate the performance of the LLMs whether this information is provided or not. Notably, for all comparative methods, this hint is used to divide the data into training and testing sets.; (2) \textbf{Separate vs Combine}: It would also be interesting to assess whether there is a significant difference in performance when presenting all the trajectories in a single prompt versus in separate prompts. This is because placing them all in one prompt might allow the model to consider interactions between different trajectories. In Table~\ref{prompt}, we present examples to illustrate the details of prompt.

\paragraph{Choices of LLMs.} We opted to utilize OpenAI's state-of-the-art models, GPT-3.5 and GPT-4, via their API system, and Claude-2 by Anthropic. Specifically, we use gpt-3.5-turbo-16k-0613 and gpt-4-0613 for `Separate' prompt, and Claude-2 for `Combine' prompt due to its capability to hold long input prompt up to 100K input context window size.

\begin{itemize}[leftmargin=*]
    \item \textbf{LLMs can effectively detect anomaly behaviors without any indicative information.} We observed that the LLM demonstrates commendable detection results on both datasets. For the Geolife dataset, Claude-2 surpasses all non-deep learning methods, achieving performance on par with the deep learning method. Both GPT-3.5 and GPT-4 also produce results that are comparable to those of other methods. This might suggest that presenting all mobility trajectories in a single prompt may lead to better performance than using separate prompts. As for the PoL dataset, the GPT-3.5 model significantly outperforms all the methods it was compared against.
    \item \textbf{Providing additional indicative information can further enhance the detection efficacy of LLMs.} We observed that by incorporating a `hint' into the LLMs, detection performance consistently improved across all models when tested on the Geolife dataset. Notably, Claude-2-with-hint demonstrated a significantly superior detection rate, surpassing all other comparison methods. On the other hand, there was a slight dip in performance on the PoL dataset when adding the hint. This could arguably be attributed to the LLM's ability to manage longer input temporal trajectories, as evidenced by the average length of trajectories being 52.2 for Geolife and 182.0 for PoL.
    \item \textbf{LLMs is capable to provide reasonable explanation to their judgements.} Examples of generated explanations alongside predictions can be found in Appendix~\ref{sec:output}. Notably, we observed that the LLMs are capable of providing cogent explanations for their prediction results. Such clarity is pivotal for ensuring transparency in anomaly detection methods.
\end{itemize}

\begin{table}[t]
\centering
\begin{adjustbox}{width=1.\columnwidth,center}
\begin{tabular}{@{}lcccccccc@{}}
\toprule
 & \multicolumn{4}{c}{Geolife} & \multicolumn{4}{c}{Patterns-of-Life} \\
\cmidrule(lr){2-5} \cmidrule(l){6-9}
Model & Top-10 Hits & Top-25 Hits\textsuperscript{*} & AP score & AUC score & Top-10 Hits & Top-100 Hits & AP score & AUC score \\
\midrule
OMPAD & 1 & 4 & 0.1665 & 0.1697 & 0 & 0 & 0.0079 & 0.4512 \\
MoNav-TT & 0 & 7 & 0.2849 & 0.3989 & 0 & 0 & 0.0094 & 0.4798 \\
TRAOD & 4 & 7 & 0.1060& 0.5498 & 0 & 1 & 0.0030 & 0.4390 \\
DSVDD & 7 & 15 & 0.6246 & 0.7714 & 1 & 2 & 0.0120 & 0.5398 \\
DAE & 5 & 12 & 0.4627 & 0.6234 & 0 & 1 & 0.0089 & 0.4649 \\
\midrule
GPT-3.5 & 5 & 8 & 0.4014 & 0.4979 & 0 & 6 & 0.0365 &  0.7572 \\
GPT-3.5-with-hint & 4 & 12 & 0.3741 & 0.5917  & 0 & 2 & 0.0176 & 0.6220 \\\hdashline
GPT-4 & 3 & 9 & 0.2732 & 0.4417 & - & - & - & - \\
GPT-4-with-hint & 5 & 8 & 0.3181 & 0.4818 & - & - & - & - \\\hdashline
Claude-2 & 4 & 13 & 0.4756 & 0.7474 & - & - & - & - \\
Claude-2-with-hint & 7 & 16 & 0.6879 & 0.8875 & - & - & - & - \\
\bottomrule
\end{tabular}

\end{adjustbox}
    \vspace{-1mm}
    \caption{Outlier detection performance for all datasets. The best performance for AP and AUC scores is highlighted for each dataset.  \textsuperscript{*}We report Top-25 Hits instead of Top-100 for Geolife dataset due to their size constraints on datasets. (-) denotes the absence of experiments due to the API cost issue. \vspace{-0.4cm}}
    \label{table:performance} 
    \vspace{-3mm}
\end{table}


\section{Conclusion and Future Works}
In this work, we conduct empirical studies to provide insights on the strengths and limitations of large language models (LLMs) for detecting anomalous behaviors from mobility data, by comparing LLMs to specialized anomaly detection methods. Our key findings show that LLMs can achieve promising anomaly detection performance even without any specific cues about potential anomalies. Furthermore, providing contextual information about possible irregularities can enhance the prediction accuracy of LLMs. In addition, LLMs can provide explanations for their anomaly judgments, thereby improving model transparency. For future work, we plan to study the effectiveness of open source LLMs such as Llama-2 models to improve model transparency. We also aim to address the issue that LLMs have difficulty processing long mobility trajectories due to the limited context window size. Moreover, we intend to evaluate our approach on additional mobility datasets. This work represents an initial exploration of applying LLMs for the important and promising task of mobility anomaly detection. We hope it will inspire more research in this direction.

\bibliographystyle{plain}
\bibliography{custom}
\appendix
\section{Comparison Methods}\label{sec:compare}
 \textbf{OMPAD} \cite{basharat2008learning} is an outlier detection method that analyzes objects' movement patterns by counting the types of locations they visit. It identifies abnormal activities by measuring the deviations in moving trends compared to established normal patterns.\\
 \textbf{MoNav-TT }\cite{zhang2012smarter} is an outlier detection algorithm tailored for urban human trajectory networks, where it detects outliers by measuring discrepancies in traffic distances. In particular, a user is identified as an outlier if the traveled distance significantly deviates from their previous behavior.\\
 \textbf{TRAOD } \cite{lee2008trajectory} is a partition-and-detect
framework for trajectory outlier detection, which partitions a
trajectory into a set of line segments, and then, detects outlying
line segments for trajectory outliers.\\
 \textbf{DSVDD} \cite{ruff2018deep} is a deep one-class classification based outlier detection method. We generalize it to handle the task of semantic trajectory outlier detection in a most intuitive way. We map the weekly trajectories of each user to a high dimensional sphere by a deep neural network encoder. Then the distance of trajectories from the sphere's surface is quantified as an outlier score.\\
 \textbf{DAE} \cite{zhou2017anomaly,dotti2020hierarchical} is a widely-used outlier detection method that leverages a deep autoencoder. Utilizing an encoder-decoder model architecture, it reconstructs input trajectories, and the resulting reconstruction error is used as an outlier indicator, signifying deviations from the normal pattern.

For the data preprocessing, we transform the semantic types of locations into onehot encodings and treat the spatial coordinates as 3D vectors as the input to all the comparison methods.

\newpage
 \section{Example Generations}\label{sec:output}
 \begin{table}[h] \centering
\begin{tabular}{p{1.5cm}p{11.5cm}}\toprule
Model & Output\\ \midrule
\textbf{GPT-3.5} \newline No hint &  Based on the given trajectory data, the anomaly score indicates the abnormal behavior of the user's mobility pattern. The analysis of the trajectory suggests that the user's movements appear highly irregular and inconsistent. The user tends to visit different types of locations (e.g., Pub, Workplace, Apartment) frequently and travels long distances within short time durations. Additionally, the user exhibits sudden changes in visited locations, such as back-and-forth movements between the same places.Furthermore, the user shows unusual patterns of movement, such as traveling long distances in short periods (e.g., a trip of 250.1 km followed by a trip of 17.4 km) or visiting distant locations in rapid succession (e.g., traveling 90.3 km to a Pub). These patterns deviate significantly from regular human mobility behavior.Considering all these factors, the estimated anomaly score for this user's behavior is [0.85]. This high score suggests that the user's mobility trajectory exhibits highly abnormal patterns and deviates significantly from typical human movements. \\
 \hline
\textbf{GPT-4} \newline No hint &   Analyzing the user's given mobility trajectory, several patterns and potential anomalies are noted. Most noticeably, the user frequents pubs substantially more than any other location type. This is somewhat unusual as typically individuals might frequently visited places like apartments or workplaces. Another significant observation is the abrupt and sporadic changes in distance to the next location - with some movements being in the order of hundreds of kilometers, indicating long-distance travel, and then returning immediately to local sites in a short span of time. The routine visits to the pubs often happen at atypical time slots, including times when a standard pub would potentially not be open or the individual would traditionally be at the workplace. The lack of visit to other essential places like shops or eateries also is a little abnormal. However, the data set does not provide any benchmark for what could be considered 'normal' behavior, making it challenging to give a precise anomaly score. Therefore, based on the noticeable inconsistencies in the temporal and spatial patterns, I would estimate the anomaly score to be moderately high, but not extreme.Estimated Anomaly Score: [0.65] \\
\bottomrule
\end{tabular}
\caption{Example LLMs' generated answers in geolife dataset without hint.}
\end{table}

\newpage
 \begin{table}[h] \centering
\begin{tabular}{p{1.5cm}p{11.5cm}}\toprule
Model & Output\\ \midrule
\textbf{GPT-3.5} \newline With hint &  Analysis and Estimated Anomaly Score: Looking at the given trajectory data, we can observe the following patterns:1. The user frequently visits the Pub and Apartment locations.2. The distances traveled between locations are varying, with some long distances and some short distances. 3. There is a ``***<deviate-point>***'' denoting a time point where the user's mobility behavior significantly changes.Based on these patterns, we can analyze the trajectory and estimate the anomaly score. Before the deviate point:- The user visits the Pub and Apartment locations alternately with varying distances.- The trajectory seems consistent and follows a pattern.After the deviate point:- The user visits the Apartment again, but the distance traveled is shorter compared to previous visits. The user then visits the Workplace, which is a new location not seen before in the trajectory.- The user visits the Pub again but travels a longer distance than before.This change in the mobility pattern, the addition of a new location, and the differences in distance traveled suggest a deviation from the user's past behavior.Based on these observations, the estimated anomaly score for this user's behavior would be relatively high. Given that larger values indicate more abnormal behavior, the estimated anomaly score can be represented as [0.9]. \\
 \hline
\textbf{GPT-4} \newline With hint &   Analyzing the trajectory data before the ``***<deviate-point>***'', the individual had a somewhat consistent pattern of movement. On weekends (Saturday), they frequently visited a Pub and their Apartment. Sometimes, they also visited a Restaurant. There was also consistent movement between their Apartment on Wednesdays and Thursdays.However, after the deviate point, there has been a significant change in behaviour. For example, they visit their Apartment on different days (Wednesday and Thursday), with a visit to the Workplace added into the mix. Moreover, they started going to the Pub also on Tuesday, which was not previously observed.The traveled distances also have increased after the deviate point showing that the user has significantly changed their mobility pattern, particularly regarding the type and average distance of trips. Overall, the sudden deviation from their previous behavioral pattern does indeed make the user suspicious, which we are considering as an anomaly. It clearly indicates that the user's activity after the ``deviate-point'' is significantly different from their past behaviours. Anomaly score is subjectively judged based on the provided context and may vary based on specific models used, but based on my analysis, I'd put it at around [0.7] as there's a clear and substantial change in behaviour, but this still has room for more drastic changes, such as the inclusion of entirely new location types or far greater distances. So it does pose a certain level of abnormality, but there's potential for it to be higher. \\
\bottomrule
\end{tabular}
\caption{Example LLMs' generated answers in geolife dataset with hint.}
\end{table}

\newpage
 \section{Prompt Example}\label{sec:prompt}

 \begin{table}[h] \centering
\begin{tabular}{p{1.5cm}p{11.5cm}}\toprule
Task \newline setting & Prompt to LLM\\ \midrule
\textbf{Separate} &  Task: You are a human mobility trajectory behavior anomaly detector. Given a historical human trajectory information, can you analyse the pattern behind the trajectory and give an anomaly score (from 0 to 1, where larger value indicates more abnormal) of this user's behavior? \newline Description of input trajectory data: A temporal sequence of visited place points, each place is consisted of the visited timestamp and its type of location. Then the traveled distance to next location is given. \newline Here is the sequence of trajector: Sat 10:36, Pub, 0.4 km ->Sat 10:57, Apartment, 4.8 km ->Sat 13:44, Pub, 3.4 km ->Sat 04:45, Apartment, 4.9 km ->Sat 07:39, Pub, 0.6 km ->Sat 09:44, Apartment, 0.8 km ->Sat 10:14, Apartment, 3.1 km ->Wed 00:30, Apartment, 1.1 km ->Wed 06:08, Apartment, 0.2 km ->Wed 06:44, Apartment, 1.3 km ->Wed 07:27, Apartment, 13.5 km ->Sat 03:39, Apartment, 0.3 km ->Sat 04:17, Apartment, 2.1 km ->Sat 05:52, Apartment, 3.7 km ->Sat 08:48, Pub, 0.6 km ->Sat 09:36, Restaurant, 11.9 km ->Thu 06:01, Apartment, 1.5 km ->Wed 08:25, Apartment, 1.3 km ->Thu 01:05, Workplace, 0.6 km ->Sat 01:18, Pub, 10.4 km ->Sat 05:33, Pub. \newline Give your analysis and present your esimated anomaly score (from 0 to 1, where larger value indicates more abnormal) inside a pair of square brackets [] : \\
 \hline
\textbf{Separate With Hint} &   Task: You are a human mobility trajectory behavior anomaly detector. Given a historical human trajectory information, can you analyse the pattern behind the trajectory and give an anomaly score (from 0 to 1, where larger value indicates more abnormal) of this user's behavior?
Hint: The anomaly users would suddenly change their mobility pattern starting from a time point, which means after a certain time, their mobility behavior would significantly deviate from their past behaviors. We would use "***<deviate-point>***" inside each trajectory to denote the time point as hint.\newline Description of input trajectory data: A temporal sequence of visited place points, each place is consisted of the visited timestamp and its type of location. Then the traveled distance to next location is given.\newline Here is the sequence of trajector: Sat 10:36, Pub, 0.4 km ->Sat 10:57, Apartment, 4.8 km ->Sat 13:44, Pub, 3.4 km ->Sat 04:45, Apartment, 4.9 km ->Sat 07:39, Pub, 0.6 km ->Sat 09:44, Apartment, 0.8 km ->Sat 10:14, Apartment, 3.1 km ->Wed 00:30, Apartment, 1.1 km ->Wed 06:08, Apartment, 0.2 km ->Wed 06:44, Apartment, 1.3 km ->Wed 07:27, Apartment, 13.5 km ->Sat 03:39, Apartment, 0.3 km ->Sat 04:17, Apartment, 2.1 km ->Sat 05:52, Apartment, 3.7 km ->Sat 08:48, Pub, 0.6 km ->Sat 09:36, Restaurant, 11.9 km ->Thu 06:01, Apartment ***<deviate-point>*** , 1.5 km ->Wed 08:25, Apartment, 1.3 km ->Thu 01:05, Workplace, 0.6 km ->Sat 01:18, Pub, 10.4 km ->Sat 05:33, Pub.\newline Give your analysis and present your esimated anomaly score (from 0 to 1, where larger value indicates more abnormal) inside a pair of square brackets [] :  \\
\bottomrule
\end{tabular}
\caption{A detailed `Separate' prompt example in geolife dataset.}
\end{table}

\end{document}